\documentclass[10pt,twocolumn,letterpaper]{article}

\usepackage[pagenumbers]{cvpr} % To force page numbers, e.g. for an arXiv version

\definecolor{cvprblue}{rgb}{0.21,0.49,0.74}
\usepackage[pagebackref,breaklinks,colorlinks,allcolors=cvprblue]{hyperref}
\usepackage{float}
\usepackage[table]{xcolor}
\usepackage{booktabs,multirow,makecell}
\usepackage{pifont} 
\usepackage{array}
\usepackage{siunitx}

\newcommand{\cmark}{\textcolor{teal}{\ding{51}}}   % 勾（绿色/青绿色）
\newcommand{\xmark}{\textcolor{red}{\ding{55}}} 
\def\paperID{} % *** Enter the Paper ID here
\def\confName{CVPR}
\def\confYear{2026}

\makeatletter
\newcommand\blfootnote[1]{%
  \begingroup
    \renewcommand\thefootnote{}% 不显示脚注符号
    \footnote{#1}% 插入脚注文本
    \addtocounter{footnote}{-1}% 恢复脚注计数
  \endgroup
}
\makeatother
\title{GraspLDP: Towards Generalizable Grasping Policy via Latent Diffusion}

\author{%
  Enda Xiang\textsuperscript{1,2},
  Haoxiang Ma\textsuperscript{1,2},
  Xinzhu Ma\textsuperscript{2},
  Zicheng Liu\textsuperscript{2},
  Di Huang\textsuperscript{1,2\textdagger}\\[6pt]
  \textsuperscript{1}State Key Laboratory of Complex and Critical Software Environment,\\ Beihang University, Beijing, 100191, China\\
  \textsuperscript{2}School of Computer Science and Engineering, Beihang University, Beijing, 100191, China\\
  {\tt\small {\{endaxiang, mahaoxiang822, xinzhuma, liuzicheng, dhuang\}@buaa.edu.cn}}
  \vspace{-14pt}
}
\begin{document}
\vspace{-14pt}
%% teaser 
\twocolumn[{% 
\renewcommand\twocolumn[1][]{#1}% 
\maketitle 
\begin{center} 
\centering 
\includegraphics[width=0.98\textwidth]{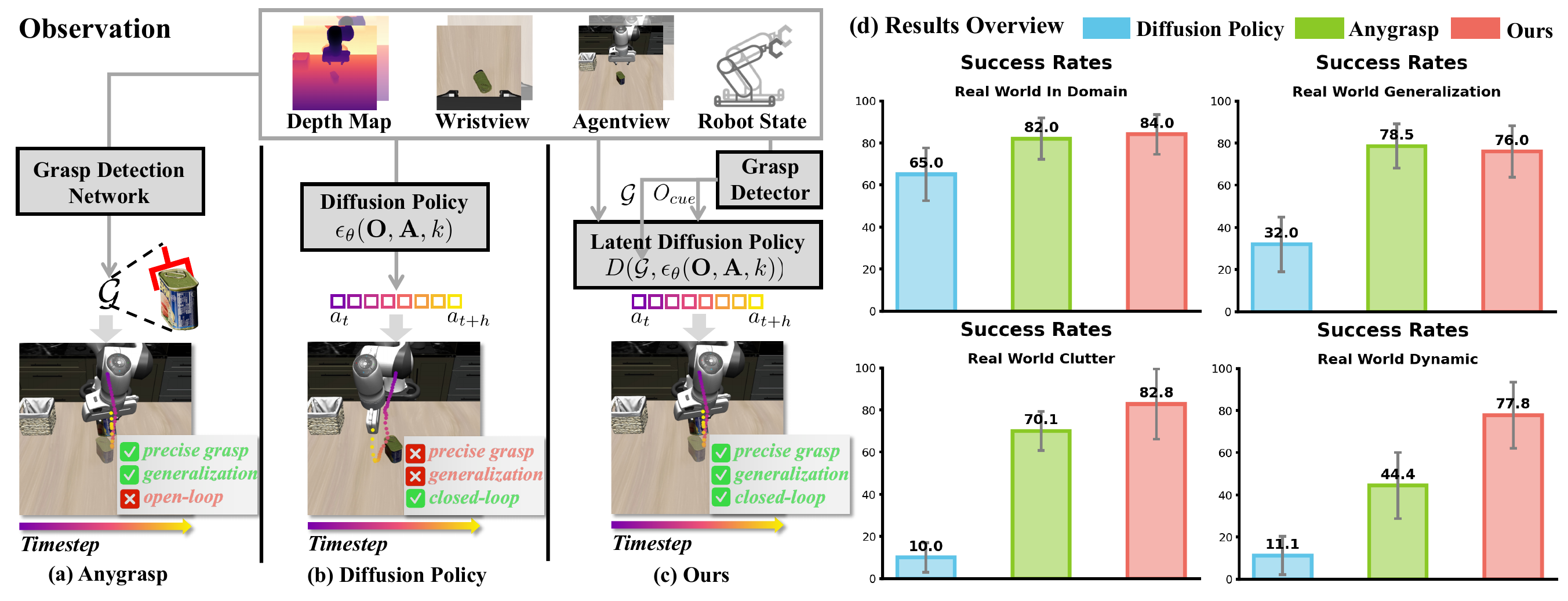} 
\captionof{figure}{We introduce \textbf{GraspLDP}, a generalizable grasping policy integrated with the prior from grasp detector via latent diffusion. Specifically, prior works generally ({\it a}) predict the grasp pose (\eg Anygrasp~\cite{fang23anygrasp}) or ({\it b}) generate action sequence (\eg Diffusion Policy~\cite{DBLP:conf/rss/ChiFDXCBS23}) for grasping. In contrast, ({\it c}) our method extracts grasp priors from a pre-trained grasp detector for action refinement in latent space, and ({\it d}) achieves substantial advantages over previous works in diverse grasping tasks. %\xz{image format: *.png to *.pdf}
} 
\label{fig:overview}
\end{center}% 
}]

\blfootnote{ \textdagger\ Corresponding Author.}
\label{sec:intro}
\begin{abstract}

This paper focuses on enhancing the grasping precision and generalization of manipulation policies learned via imitation learning. Diffusion-based policy learning methods have recently become the mainstream approach for robotic manipulation tasks. As grasping is a critical subtask in manipulation, the ability of imitation-learned policies to execute precise and generalizable grasps merits particular attention. Existing imitation learning techniques for grasping often suffer from imprecise grasp executions, limited spatial generalization, and poor object generalization. To address these challenges, we incorporate grasp prior knowledge into the diffusion policy framework. In particular, we employ a latent diffusion policy to guide action chunk decoding with grasp pose prior, ensuring that generated motion trajectories adhere closely to feasible grasp configurations. Furthermore, we introduce a self-supervised reconstruction objective during diffusion to embed the graspness prior: at each reverse diffusion step, we reconstruct wrist-camera images back-projected the graspness from the intermediate representations. Both simulation and real robot experiments demonstrate that our approach significantly outperforms baseline methods and exhibits strong dynamic grasping capabilities. Project Page can be found \href{https://coolmakersss.github.io/GraspLDP.github.io/}{here}.
%\xz{Both simulation and real robot experiments demonstrate that our approach significantly outperforms baseline methods by achiveing \textbf{84.0\%}, \textbf{76.0\%} and \textbf{82.8\%} success rate in different settings and exhibits strong dynamic grasping capabilities with success rate of \textbf{77.8\%}.}

\end{abstract}    

\section{Introduction}
\label{sec:intro}
% Recently, visual imitation learning has experienced rapid development, and a spectrum of visuomotor policies\cite{DBLP:conf/rss/ZhaoKLF23, DBLP:conf/rss/ChiFDXCBS23, DBLP:conf/corl/KimPKXB0RFSVKBT24, black2410pi0} has achieved remarkable results on robotic manipulation tasks. Building on this progress, numerous studies have sought to improve the action precision and generalization of such policies, yielding notable advances. In this work we focus on grasping, a core and foundational manipulation skill. We observe that for grasping or pick-and-place tasks, conventional policies such as vanilla diffusion policy\cite{DBLP:conf/rss/ChiFDXCBS23} can still fail due to inaccurate end-effector poses as illustrated in Figure \ref{fig:overview} (a). Furthermore, when the spatial pose of an object varies or novel objects are introduced, limited spatial generalization and poor object generalization often produce sharp declines in task success rates.

Within the broader robotic manipulation pipeline, grasping serves as a pivotal initial step that enables physical interaction. Besides, grasp detection methods have achieved remarkable results by mapping visual inputs, such as images or point-clouds, to viable grasp poses. Recently, visuomotor policies derived from imitation learning~\cite{DBLP:conf/rss/ZhaoKLF23, DBLP:conf/rss/ChiFDXCBS23, DBLP:conf/corl/KimPKXB0RFSVKBT24, black2410pi0} have demonstrated significant potential in general-purpose robotic manipulation. Trained on large-scale demonstration data, these policies can exhibit zero-shot generalization and achieve robust error correction against external disturbances via closed-loop control. However, specifically for the grasping stage, these general-purpose policies often fall short of specialized grasp detection methods, as modeling the entire action sequence of grasping is an inherently more complex task. Therefore, enhancing the capability of general visuomotor policies to perform fine-grained generalized grasping is a critical research objective.

To address this issue, prior research has primarily focused on two directions: data-centric strategies and the integration of additional prior knowledge. From the data-centric perspective, to alleviate the scarcity of grasping demonstrations, GraspVLA~\cite{graspvla} generates a massive dataset Syngrasp-1b of 1 billion simulated frames to train a Vision-Language-Action (VLA) model, which in turn demonstrates remarkable zero-shot sim-to-real transfer capabilities. However, large-scale data generation incurs substantial computational cost: Syngrasp-1b consumes 160 RTX 4090 GPUs for 10 days to simulate, an really uneconomical expense. On the other hand, large VLA models also suffer from high inference latency and low action frequency, which hinders real-time grasping in dynamic scenes. Regarding the integration of additional knowledge, some works~\cite{robograsp, SpatialRoboGrasp,gpa-ram} enhance imitation learning frameworks to enhance grasp performance by incorporating a grasp detection network, whose accurate predictions of the target grasp pose provide efficient guidance for the policy model. Compared to GraspVLA, these methods can achieve more efficient usage of the demonstration data and reduce inference latency. However, these methods treat grasp poses merely as conditional input for the policy model, which leads to two issues. On the one hand, the input grasp pose is weakly correlated with the output action sequence, making it difficult to provide efficient guidance. On the other hand, the mismatch between the low-semantic grasp pose and the visual inputs causes the policy model to fail to adequately extract information about the spatial distribution of grasps.

%The overlook of mismatch between low-semantic grasp pose information and visual features cause the policy model fails to adequately correlate the input grasp pose with the output action sequence, resulting in a suboptimal use of the grasp prior.

% They also don't thoroughly study how grasp poses can guide trajectories or design imitation frameworks that adapt to such guidance.
%加SD3引用
To overcome these challenges, drawing inspiration from recent advances in image generation~\cite{avrahami2023blended,zeng2024hairdiffusion,wu2025latentps}, we introduce a novel grasp guidance framework built upon latent diffusion models. In contrast to previous methods, our key innovation is to steer the action generation process by constructing an action latent space and injecting the precise target grasp pose. As illustrated in Fig.~\ref{fig:overview} (c). Instead of holistically modeling the entire grasp sequence with a single policy, our method disentangles the action latent generation into two distinct components: a target grasp pose and the corresponding motion policy. The former is predicted by a dedicated grasp detection network, while the latter is learned by the diffusion model. This decomposition bridges the gap between the static target grasp pose and the dynamic action sequence by projecting both into a shared latent space. Furthermore, to minimize the mismatch between grasp pose and visual inputs, we provide the latent diffusion policy with a visual graspness cue. This cue is a graspness map~\cite{Graspness}, which is also generated by the grasp detector, explicitly directs the policy's attention toward grasp-relevant regions.

% novel visuomotor policy that integrates a pre-trained grasp detector to improve grasping performance. We extract two types of priors from the detector: the
%  graspness and grasp pose prior, corresponding to “where to grasp” and “how to grasp” respectively\cite{Graspness}. The graspness prior denotes geometry-based graspable regions; we back-project this prior into wrist-view images to provide reliable visual cues for end-effector motion and enhance it via self-supervised reconstruction. For grasp poses, we design a latent diffusion policy which is inspired by conditional image editing with latent diffusion models\cite{avrahami2023blended,zeng2024hairdiffusion,wu2025latentps} in image editing field.\cite{avrahami2023blended} supplying a mask or structural condition steers the diffusion process to alter a localized region while preserving the surrounding content. Similarly, if we transfer image space into action space, faster denoise process and better action trajectory consistency is also significant in close-loop control of robotic manipulation and grasp.
 %
 %grasp poses are combined with denoised action latent from the diffusion policy and jointly decoded to produce action chunking, thereby enabling grasp-pose-conditioned guidance for training and generation of action trajectories.
 %feeding a precise grasp pose as a conditioning signal steers the action-generation process to produce motion that realizes a targeted grasp while preserving broader trajectory consistency.
In this paper, we propose a framework for generalizable grasping policy via latent diffusion following~\cite{rombach2022high}. Under this two-stage framework, we effectively integrate priors including graspness map and grasp pose from grasp detection network. For grasp pose prior, we refine action chunks under the guidance of a grasp pose in latent space encoded by a Variational Auto-Encoder (VAE) in Action Latent Learning stage, enabling more effective steering of low-semantic information. For graspness map prior, we attach the graspness map to the wrist camera image as a visual cue to condition the diffusion model’s denoising process in Diffusion on Latent Action Space stage. In each denoising step, we reconstruct the wrist-view image as an auxiliary self-supervised objective to strengthen the policy’s conditioning on the graspness cue. During inference, we further propose Heuristic Pose Selector (HPS) which jointly considers grasp pose quality and the current end-effector state to choose the most appropriate grasp pose from candidates as guidance. Experimental results show that our method further improves in-domain grasping success rate by \textbf{17.5\%} compared to diffusion policy, and yields significant gains in spatial, object and visual generalization of \textbf{22.2\%}, \textbf{46.8\%} and \textbf{48.3\%}, respectively. We also find that our approach remains highly effective in dynamic grasping, demonstrating the practical potential of the proposed method.

\section{Related work}
\label{sec:related_work}

%\subsection{Grasp Detection}
\noindent \textbf{Grasp Detection.} Grasp detection has been extensively studied over the past decade. This task typically aims to predict feasible grasp poses for objects based on visual observations. GPD~\cite{ten2017grasp} pioneers a two-stage pipeline that combined a sampling algorithm in large-scale candidates with a CNN-based scoring module, achieving high precision in 6-DoF grsap detection. Subsequently, end-to-end grasp detection methods~\cite{mousavian20196, fang2020graspnet, Graspness, ma2023towards, fang23anygrasp, ma2024sim, ma2024generalizing} become a major research focus.~\cite{mousavian20196} proposes a conditional variational grasp generator that models multimodal 6-DoF grasp distributions. GraspNet-1Billion~\cite{fang2020graspnet} is a large-scale benchmark for general grasping that contains over a billion labeled grasp candidates, greatly alleviating data scarcity and evaluation inconsistency. GSNet~\cite{Graspness} introduces the notion of graspness measures the point-wise graspability in the point-cloud. AnyGrasp~\cite{fang23anygrasp} addresses the spatiotemporal dimension with dense supervision and demonstrates human-like performance on bin-picking and dynamic grasping tasks. With the rise of LLMs and VLMs, works like GraspGPT~\cite{tang2023graspgpt} and ThinkGrasp~\cite{qian2025thinkgrasp} have begun to integrate vision–language reasoning to enable task-oriented grasp detection in cluttered scenes. While these open-loop grasp approaches achieve respectable accuracy, the absence of closed-loop perception during grasp execution limits their adaptability and degrades performance in dynamic environments.

%\subsection{Visual Imitation Learning}
\noindent \textbf{Visual Imitation Learning.} In recent years, visual imitation learning has advanced substantially. Early behavior cloning works~\cite{florence2022implicit, jang2022bc} fitted mappings from robot states to actions using expert demonstrations to accomplish robotic manipulation tasks. Subsequent methods such as ACT~\cite{DBLP:conf/rss/ZhaoKLF23} and Diffusion Policy~\cite{DBLP:conf/rss/ChiFDXCBS23} leverage generative models to produce action chunk from 2D visual observations, achieving significant success. Following~\cite{DBLP:conf/rss/ChiFDXCBS23}, diffusion-based models became widely adopted for visual imitation learning. 3D Diffusion Policy~\cite{ze20243d} and 3D Diffuser Actor~\cite{ke20243d} extend 2D visual observations to 3D space to enhance spatial perception.~\cite{wang2024gendp, chen2025g3flow} and ~\cite{wang2025equivariant, tie2025seed} introduce 3D semantic fields and equivariance into diffusion policies respectively to improve sample efficiency and generalization. RDP~\cite{xue2025reactive} proposes a visual-tactile diffusion framework that integrates tactile or force information for contact-rich manipulation task. Meanwhile, large VLA models have also begun to adopt diffusion head for action generation after preliminary explorations\cite{zitkovich2023rt, DBLP:conf/corl/KimPKXB0RFSVKBT24}. Octo~\cite{ghosh2024octo} uses a diffusion action head to predict action chunking. Then RDT-1B~\cite{liu2024rdt} builds a much larger DiT~\cite{peebles2023scalable} to realize general bimanual manipulation. $\pi_0$~\cite{black2410pi0} and $\pi_{0.5}$~\cite{intelligence2025pi_} employs flow-matching~\cite{lipmanflow} to achieve higher-frequency and smoother action sequence. Due to the absence of task-specific design for grasping, policy-based approaches are often suboptimal. In contrast, our method effectively addresses the limitations of closed-loop grasping strategies in terms of accuracy and generalization by incorporating prior knowledge of grasping.

%\subsection{Visuomotor Policy for Grasping}
\noindent \textbf{Visuomotor Policy for Grasping.}
In this section, we focus on works that design imitation learning policies for grasping. GraspVLA~\cite{graspvla} is a data-centric approach for universal grasping. Fueled by one billion frames of data, it explicitly adopts a Chain-of-Thought (CoT)~\cite{wei2022chain} pipeline named Progressive Action Generation, causing the model to output the bounding box and the grasp pose of the target object before generating the action chunk by flow matching action expert. This pipeline achieves state-of-the-art results in universal closed-loop grasping. Simultaneously, some prior-centric approaches have been proposed that integrate grasp detection module or grasp pose into imitation learning workflows. PPI~\cite{yang2025gripper} first leverages discrete key poses (which can be viewed as generalized grasp poses) to guide continuous action generation: the paper treats object pointflow and key gripper poses as intermediate interfaces that steer denoising process, showing effectiveness on long-horizon bimanual tasks. Robograsp~\cite{SpatialRoboGrasp} injects pose-aware grasping features for planar grasps as one conditioning input to a diffusion policy and synchronously calls a pre-trained grasp detector at every inference timestep; Spatial Robograsp~\cite{robograsp} extends this idea to full 6-DoF grasps. GPA-RAM~\cite{gpa-ram} uses pre-trained M2T2~\cite{yuan2023m2t2} and ResNet as grasp detectors and fuses their implicit features with Spatial Attention Mamba output tokens to predict the next key pose for task execution. In summary, current research has not fully mined the mature grasp detection literature for richer priors---most efforts remain limited to grasp poses---and the methods for introducing those priors still require further exploration.

\section{Method}
\label{sec:method}

%\begin{figure*}[h!]
%  \centering
%  \includegraphics[width=0.95\textwidth]{fig/framework.png}
%  \caption{Framework of proposed GraspLDP. Stage 1 and Stage 2 describe our two stage train pipeline. Inference Pre-process presents our inference pipeline with pre-trained grasp detector.}
%  \label{fig:framework}
%\end{figure*}

\subsection{Overview}
\begin{figure*}[h!]
  \centering
  \includegraphics[width=0.90\textwidth]{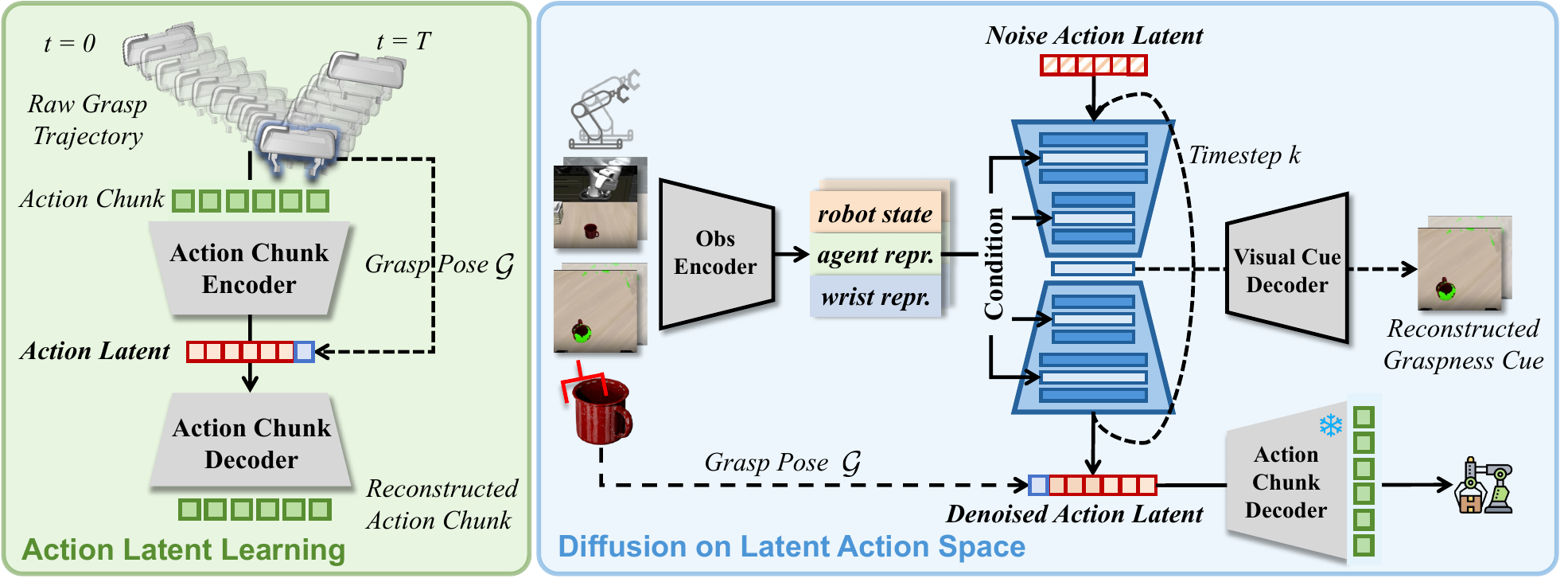}
  \caption{Framework of proposed GraspLDP. In Action Latent Learning stage action chunks are refined under the guidance of a grasp pose in latent space encoded by a VAE.  In Diffusion on Latent Action Space stage the graspness cue is used to condition the diffusion model’s denoising process and to reconstruct for enhancement.}
  \label{fig:framework1}
\end{figure*}

Our method can be formulated as learning a visuomotor policy $\pi: \mathbf{\mathcal{O}} \rightarrow \mathcal{A}$ that maps observations to action chunks. Our core insight is that grasp priors can not only increase the accuracy of actions in grasping tasks by providing precise grasp configurations and improve generalization by reducing the policy’s reliance on raw visual observations and proprioceptive states. To this end, we propose \textbf{GraspLDP}, a two-stage trained latent diffusion model as shown in Figure \ref{fig:framework1}. The detailed design rationale and technical specifics are presented in the following subsections.

\subsection{Grasp Guidance in Latent Space}

Diffusion policy can be formulated as a conditional denoising model conditions on the current timestep observation $O_t$, which includes RGB images, depth maps, and the robot’s proprioceptive states. Concretely, starting from Gaussian noise $A_t^{k}$, the denoising network $\epsilon_{\theta}$ performs k steps of a Markov process with parameterized Gaussian transitions to produce the final clean action chunk $A_t^{0}$:
\begin{equation}
A_t^{k-1} = \alpha\left( A_t^{k} - \gamma\epsilon_{\theta}(O_t, A_t^{k}, k) + \mathcal{N}(0,\sigma^2 \mathbf{I})\right).
\end{equation}
To generate actions guided by a target grasp pose $\mathcal{G}$, a natural idea is to include the pose directly as part of the observation or proprioceptive state and condition the denoising process on it. On the one hand, modeling the joint conditional distribution in this way both dilutes the guiding strength of the grasp pose and makes denoising training harder. On the other hand, performing grasp detection and action-chunk denoising sequentially increases inference latency, and low policy latency is itself important for task success. Inspired by prior work~\cite{tan2024multi, xue2025reactive}, we therefore adopt a latent diffusion policy. Concretely, we first use a lightweight VAE encoder to compress action chunks into compact action latents $\mathbf{Z}$:
\begin{equation}
\mathbf{Z} = \mathcal{E}(A).
\end{equation}
These latents are paired with the trajectory’s corresponding grasp pose $\mathcal{G}$ and then reconstructed into action chunks by a asymmetric decoder:
\begin{equation}
\hat{A} = \mathcal{D}(\mathbf{Z}\oplus \mathcal{G}).
\label{eq:0}
\end{equation}
The VAE is trained using an L2 reconstruction loss with a Kullback-Leibler (KL) penalty loss~\cite{kingma2013auto} as follows:
\begin{equation}
\mathcal{L}_{VAE} = \operatorname{MSE}(A,\hat{A}) + \lambda \mathcal{L}_{KL}.
\end{equation}
Under this scheme, the denoising objective for our latent diffusion policy targets the much more compact action latent representation.

%\begin{equation}
%\mathcal{L} = %\operatorname{MSE}\!\left(\epsilon^{k},\; 
%\epsilon_{\theta}\!\big(\bar{\alpha}_{k} A^{0} + %\bar{\beta}_{k} \epsilon^{k},\; O,\;k)\right)
%\end{equation}

\subsection{Visual Graspness Cue}
\label{sec:3-3}
Graspness measures the likelihood that a point in point-clouds affords a feasible grasp. While this notion superficially resembles the concept of affordance~\cite{baldauf2010attentional, song2015learning, corona2020ganhand}, graspness is in fact a more precise, geometry-driven form of grasp affordance. As illustrated in Figure \ref{fig:framework1}, we compute point-wise graspness score $s_i\in [0,1]$ over the depth-projected point-cloud using a pretrained graspness network. 

Following the success of previous visual prompting schemes in robotic manipulation, we back-project graspness score to pixel space $\Omega$ obtaining graspness map $M$:
\begin{equation}
\Omega=\{(j,k)\mid 0\le j<H,\ 0\le k<W\},
\end{equation}
\begin{equation}
M(j,k)=s_{\pi(j,k)}, \ (j,k)\in\Omega,
\end{equation}
where $\pi: \pi(j,k) = i$ is the operator that projects pixels to 3D points.
Then the graspness map $M$ is superimposed on the wrist-view RGB image and only points whose graspness exceeds a threshold $\tau$ will be counted to mitigate excessive noise contamination to preserve the information content of the original image:
\begin{align}
O_{cue}(j,k) = 
\begin{cases}
O_{wrist}(j,k),  & M(j,k) \le \tau , \\
masked\_color, & M(j,k)>\tau
\end{cases}.
\end{align}
Then we directly use the resulting masked image as a visual cue to condition the denoising process of policy, aiming to directs the motion towards graspable regions. At the same time, to encourage the model to attend to these visual cues rather than simply rely on condition, we use $O_{cue}$ as an auxiliary self-supervised learning (SSL) objective:
\begin{equation}
\mathcal{L}_{Recon.} = \operatorname{MSE}(O_{cue},\hat{O}_{cue}),
\end{equation}
we reconstruct $O_{cue}$ from intermediate representations of the reverse diffusion process and optimize this reconstruction jointly with the diffusion loss:
\begin{equation}
\mathcal{L}_{Diff.} = \operatorname{MSE} \left(\epsilon^{k},\; 
\epsilon_{\theta} (\bar{\alpha}_{k} \mathbf{Z}^{0} + \bar{\beta}_{k} \epsilon^{k},\; O,\;k)\right),
\end{equation}
\begin{equation}
\mathcal{L}_{LDP} = \mathcal{L}_{Diff.} + \lambda_{Recon.}\mathcal{L}_{Recon.}
\end{equation}

\subsection{Heuristic Pose Selector}

\begin{figure}[h!]
  \centering
  \includegraphics[width=0.46\textwidth]{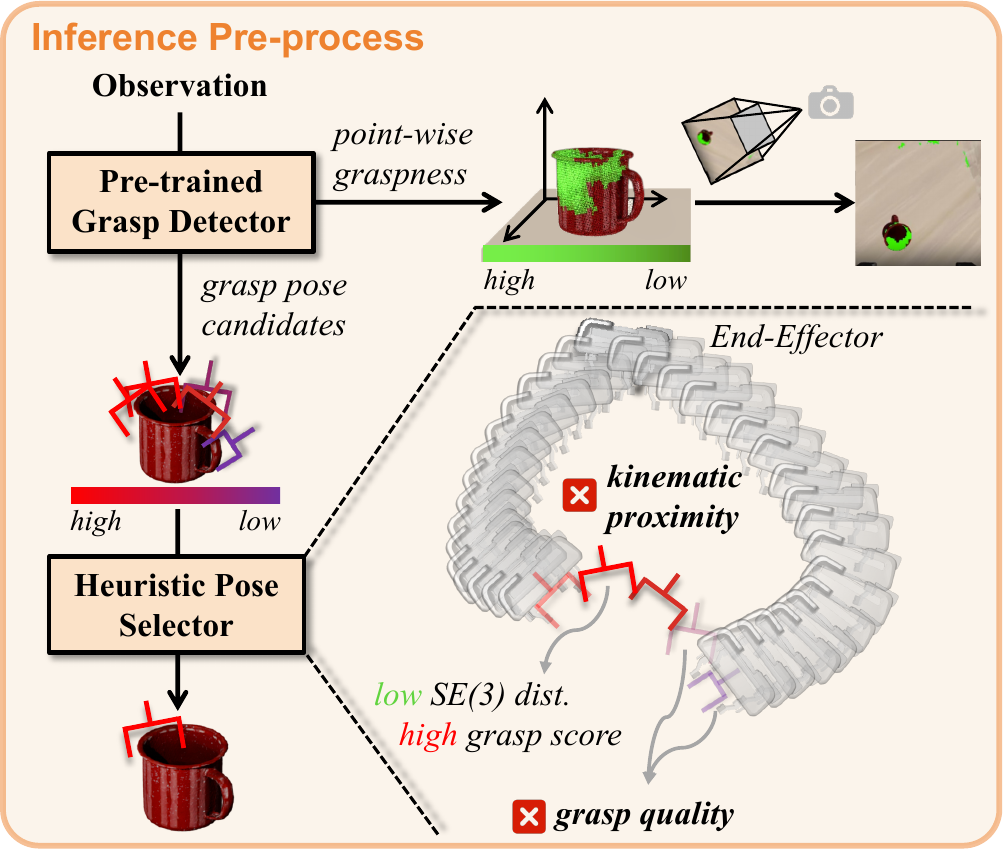}
  \caption{Inference Pre-process presents our inference pipeline with Heuristic Pose Selector.}
  \label{fig:framework2}
\end{figure}

In the inference pipeline, a key challenge is how to select an appropriate from grasp pose candidates prdiected by pre-trained grasp detector, since evidently unsuitable pose guidance could degrade the success rate during grasp execution. This process can be formally defined as selecting $\mathcal{G}^*$ from $\mathcal{G}=\{\mathcal{G}_0,\mathcal{G}_1,\dots,\mathcal{G}_m\}$. We consider two factors in the selection: the score of the grasp predicted by grasp detector and the spatial relationship between the grasp pose and the current end-effector pose. The former represents the intrinsic quality of the grasp, while the latter governs kinematic proximity, yielding smoother and more kinematically feasible end-effector trajectories. Based on these factors we implement a Heuristic Pose Selector (HPS).

First we discard any grasp poses that collide with the environment using collision detection, and then apply non-maximum suppression (NMS) to avoid highly redundant candidate poses.
We then score the remaining grasps by their grasp score which is also predicted by grasp detector and keep the top-$k$ candidates, forming the shortlist $\mathcal{G}=\{ \mathcal{G}_{m0},\mathcal{G}_{m1},\dots,\mathcal{G}_{mk}\}$.
Given the current end-effector pose $P$, we measure the pose error between $P$ and each candidate $\mathcal{G}_j$ using a pseudo-metric called SE(3) geodesic distance $d$ since there is no bi-invariant metric on SE(3) space~\cite{barfoot2014associating, blanco2021tutorial}:
\begin{equation}
\label{eq:1}
\xi=\log(P^{-1}\mathcal{G}_j),\; j \in \{m_0,m_1,\dots,m_k\},
\end{equation}
\begin{equation}
\label{eq:2}
d_{\mathcal{G}_j,W}=\sqrt{\xi^\top W\,\xi} .
\end{equation}
Here $W=\operatorname{diag}(w_t,w_t,w_t,w_r,w_r,w_r)$ is a $6\times6$ diagonal weighting matrix used to make rotation and translation commensurate. We then select the grasp $\mathcal{G}^*$ that minimizes this distance $d_j$ as our final grasp choice:
\begin{equation}
\mathcal{G}^*=\underset{\mathcal{G}_j\in \mathcal{G}}{\arg\min}\; d(\mathcal{G}_j).
\end{equation}
 With the selected grasp and denoised action latent, the VAE decoder reconstructs the final action chunk as Equation \ref{eq:0}, which will be excluded by manipulator.

\section{Experiments}
\label{sec:experiments}
\subsection{Experimental Setup}
\textbf{Benchmark.}
 We conduct simulation experiments on the LIBERO~\cite{liu2023libero} benchmark covering data collection, training, and evaluation. We curate and filter a training set of roughly \textbf{12K high-quality demonstrations} over \textbf{20 objects}, covering varied object poses and diverse grasp poses. For evaluation of generalization we select unseen objects from both the similar and novel splits in Graspnet-1billion \cite{fang2020graspnet}, which are held-out not only for our GraspLDP but also for the pre-trained grasp detection network used in the pipeline to ensure a fair comparison. More detailed procedures for constructing the benchmark are provided in the Appendix.

\begin{table*}
  \centering
  \small
  \caption{Results of evaluation in simulator. In Domain denotes cases where both the objects and their poses were present in the training data; Spatial Generalization measures how well the model handles those training objects placed in unseen poses; Object Generalization assesses performance on entirely novel objects; and Visual Generalization tests robustness under visual disturbances like lighting changes. The $\dagger$ refers to the model that has been fine-tuned on our dataset for fair comparison.}
  % 列定义：l | r@{\ (}l r@{\ (}l r@{\ (}l r@{\ (}l | c
  % 每个指标拆为两列：主值 (右对齐) 和 括号内内容 (左对齐)，@{\ (} 自动插入左括号，@{)} 自动插入右括号
  \begin{tabular}{l|
                  r@{\ }l
                  r@{\ }l
                  r@{\ }l
                  r@{\ }l
                  c}
    \toprule
    \textbf{Method}
      & \multicolumn{2}{m{2.3cm}}{\textbf{In Domain}}
      & \multicolumn{2}{m{2.3cm}}{\textbf{Spatial\newline Generalization}}
      & \multicolumn{2}{m{2.3cm}}{\textbf{Object\newline Generalization}}
      & \multicolumn{2}{m{2.3cm}}{\textbf{Visual\newline Generalization}}
      & \textbf{Average} \\
    \midrule \midrule
    Diffusion Policy~\cite{DBLP:conf/rss/ChiFDXCBS23}
      & 62.8 & (204/325)
      & 48.9 & (159/325)
      & 11.4 & (37/325)
      & 16.3 & (53/325)
      & 34.9 \\
    OpenVLA~\cite{DBLP:conf/corl/KimPKXB0RFSVKBT24}
      & 1.2  & (4/325)
      & 0.9  & (3/325)
      & 1.5  & (5/325)
      & 0.0  & (0/325)
      & 0.9 \\
    OpenVLA$\dagger$~\cite{DBLP:conf/corl/KimPKXB0RFSVKBT24}
      & 57.5 & (187/325)
      & 41.2 & (134/325)
      & 14.5 & (47/325)
      & 12.3 & (40/325)
      & 31.4 \\
    GraspVLA~\cite{graspvla}
      & 50.8 & (165/325)
      & 49.5 & (161/325)
      & 46.8 & (152/325)
      & 51.7 & (168/325)
      & 49.7 \\
    \midrule
    Ours Baseline
      & 72.3 & (235/325)
      & 59.1 & (192/325)
      & 48.3 & (157/325)
      & 47.7 & (155/325)
      & 56.9 \\
    \rowcolor{gray!20}
    \textbf{GraspLDP}
      & \multicolumn{2}{m{2.3cm}}{\textbf{80.3 (261/325)}}
      & \multicolumn{2}{m{2.3cm}}{\textbf{71.1 (231/325)}}
      & \multicolumn{2}{m{2.3cm}}{\textbf{58.2 (189/325)}}
      & \multicolumn{2}{m{2.3cm}}{\textbf{64.6 (210/325)}}
      & \textbf{68.6} \\
    \bottomrule
  \end{tabular}
  \label{tab:main}
\end{table*}

\noindent \textbf{Baselines.}
To validate the architectural advances of our GraspLDP, we compare to vanilla Diffusion Policy~\cite{DBLP:conf/rss/ChiFDXCBS23}, and to the generalist manipulation policy OpenVLA~\cite{DBLP:conf/corl/KimPKXB0RFSVKBT24}---both in zero-shot setting and in a version fine-tuned on our dataset. Since our work focuses specifically on grasping, we also compare with the state-of-the-art method GraspVLA~\cite{graspvla}. In addition, we define an Ours Baseline like prior studies~\cite{robograsp, SpatialRoboGrasp} which treats the grasp pose merely as another conditioning input concatenated with other observations during the denoising process and doesn't use our two-stage latent design.

%\noindent \textbf{Model implementations} 
%For grasp detection we use a pretrained GSNet~\cite{Graspness} (the core grasp detection module of AnyGrasp~\cite{fang23anygrasp}). Our diffusion policy uses a Unet-based diffusion reflecting that grasping is not a long-horizon sequential task. The observation decoder architecture is inspired by~\cite{li2024crossway}, a proir work that injected auxiliary self-supervised reconstruction into a diffusion policy. All rotations are represented with the 6D rotation representation~\cite{zhou2019continuity}, which provides better continuity in the numerical and 3D rotation spaces for neural network learning. For training we run the stage-1 VAE for 800 epochs to ensure high-quality action reconstruction; the stage-2 diffusion policy is trained for 500 epochs with a reconstruction weighting of $\lambda=0.2$.

\noindent \textbf{Metric.} In all the following experiments, \textbf{Success Rate (SR)} is defined as the percentage of trails that successfully complete the grasping task out of the total number of trails with the maximum time-step threshold $T_{\max}=150$. A grasp is counted as successful when the target object is stably gripped and lifted above a specified height.
For cluttered scenarios evaluation, \textbf{Scene Completion Rate (SCR)} is adopted for evaluating the percentage of how many objects of the scene successfully grasped within the allowed attempts.
To quantify how well our method follows the target grasp pose guidance $T_{GP}$, we introduce a new metric, \textbf{Grasp Frame Error (GFE)}. For each trajectory, we denote the pose of frame at which the gripper begins to close as Grasp Frame Pose (GFP). For $T_{GFP},T_{GP} \in SE(3)$ we follow Equations \eqref{eq:1} and \eqref{eq:2} and compute the GFE as defined there.
\begin{equation}
\label{eq:3}
\textbf{GFE} = \sqrt{\log(T_{GFP}^{-1}T_{GP})^\top\,W\,\log(T_{GFP}^{-1}T_{GP})}.
\end{equation}

\subsection{Simulation Evaluation}

\noindent \textbf{In Domain Evaluation.} To evaluate the capacity of our policy, we first conduct in-domain experiments where both the objects and their poses appear in the training set. For each test object we evaluate five distinct object poses and repeat each pose 13 trials, which ensures to capture the diversity of possible grasping strategies and the large number of repetitions increases statistical reliability.

\begin{figure*}[htbp!]
  \centering
  \includegraphics[width=0.96\textwidth]{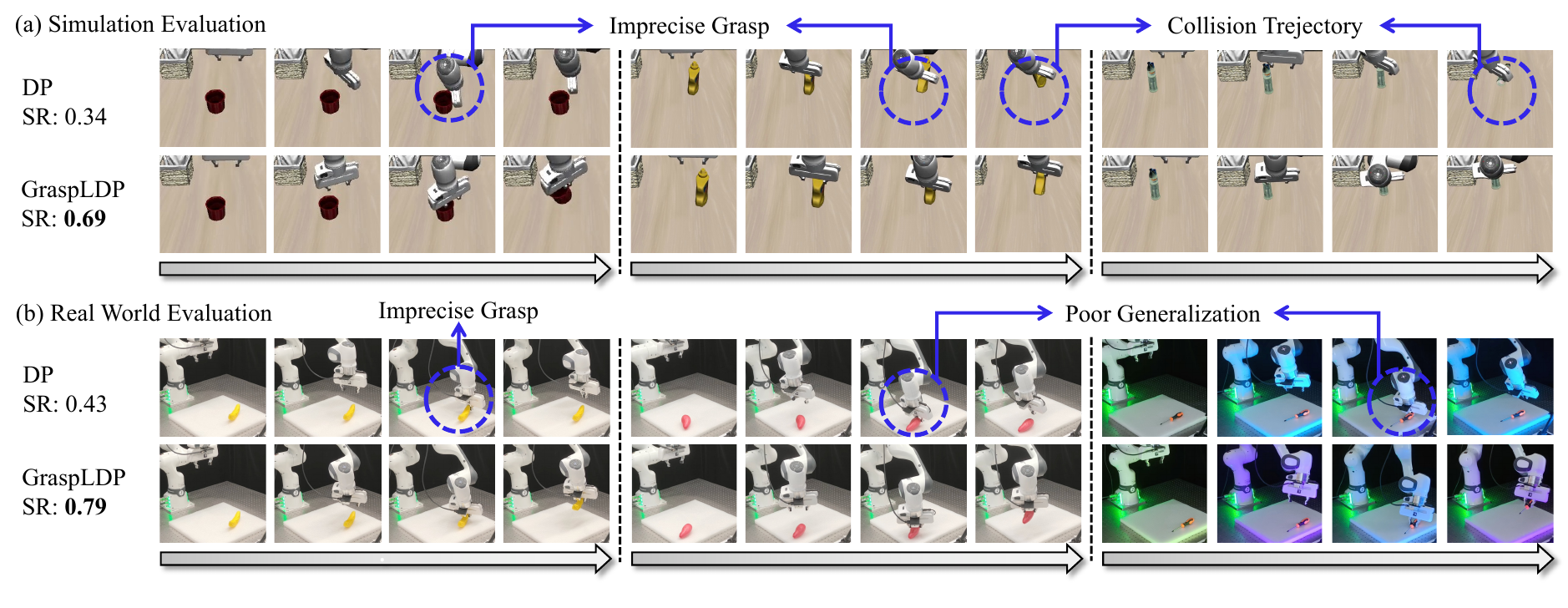}
  \caption{Qualitative experimental analysis. (a)  Grasping trials using objects "mug", "mustard bottle", and "thera med" in simulator. (b) Real world grasping trials corresponding to in domain, object generation, and visual generation performance. In particular, we use colored LED strips in low-light conditions to simulate visual interference.}
  \label{fig:vis_result}
\end{figure*}

As shown in Table \ref{tab:main}, the Diffusion Policy achieves a relatively good grasp success rate in the in-domain test set, while the fine-tuned OpenVLA still achieves only 57.5\% SR. GraspVLA demonstrates strong zero-shot grasping ability, but with high variance: it achieves nearly 100\% on some objects while nearly 0\% on others. In contrast, GraspLDP attains the highest SR of \textbf{80.3\%}. By effectively incorporating grasp priors, our policy achieves greater grasping precision as shown in Figure \ref{fig:vis_result}, whereas pervious methods often produce imprecise grasp frame poses that lead to collisions or outright grasp failures.

\begin{table*}[h!]
  \centering
  \caption{Results of ablation study. ID, SG, OG, and VG denote In Domain, Spatial, Object and Visual Generalization, respectively. GC and LG denotes Graspness Cue and Latent Guidance. CG denotes Condition Guidance used in Ours Baseline.}
  \begingroup
  \small
  {\setlength{\tabcolsep}{3pt}%
  %\renewcommand{\arraystretch}{1.5}%
  %\fontsize{9}{9}\selectfont % or \small / \scriptsize
  \begin{tabular}{m{2.0cm}m{1.6cm}m{1.6cm}m{1.6cm}m{1.6cm}m{1.6cm}m{1.6cm}m{1.6cm}m{1.6cm}@{\hskip 4pt}}
    \toprule
    \multirow{2}{*}{\textbf{Method}} & 
    \multicolumn{2}{c}{\textbf{ID}} &\multicolumn{2}{c}{\textbf{SG}} &\multicolumn{2}{c}{\textbf{OG}} &\multicolumn{2}{c}{\textbf{VG}} \\
    \cmidrule(lr){2-3}
    \cmidrule(lr){4-5}
    \cmidrule(lr){6-7}
    \cmidrule(lr){8-9}
    &
     \centering\textbf{SR$\uparrow$} & \centering\textbf{GFE$\downarrow$}  & \centering\textbf{SR$\uparrow$} & \centering\textbf{GFE$\downarrow$}  & \centering\textbf{SR$\uparrow$} & \centering\textbf{GFE$\downarrow$}  & \centering\textbf{SR$\uparrow$} & \ \ \ \ \ \ \textbf{GFE$\downarrow$} 
    \\
    \midrule \midrule
    \rowcolor{gray!20}
    \textbf{GraspLDP} &\textbf{80.3}  & \textbf{1.33}  & \textbf{71.1} & \textbf{1.61} & \textbf{58.2} & \textbf{2.10} & \textbf{64.6} & \textbf{1.92}\\
    %Ours Baseline & 72.3 & 1.97 & 57.5 & 2.48 & 48.3 & 2.68 & 47.7 & 2.75\\
    w/o GC & 77.4 \textcolor{blue!100!black}{(-2.9)} & 1.49 \textcolor{green!60!black}{(+0.16)} & 67.3 \textcolor{blue!100!black}{(-3.8)} & 1.81 \textcolor{green!60!black}{(+0.20)} & 54.2 \textcolor{blue!100!black}{(-4.0)} & 2.33 \textcolor{green!60!black}{(+0.23)} & 57.5 \textcolor{blue!100!black}{(-7.1)} & 2.30 \textcolor{green!60!black}{(+0.38)}\\
    w/o LG w/ CG & 73.5 \textcolor{blue!100!black}{(-6.8)} & 1.58 \textcolor{green!60!black}{(+0.25)} & 62.2 \textcolor{blue!100!black}{(-8.9)} & 2.07 \textcolor{green!60!black}{(+0.46)} & 52.3 \textcolor{blue!100!black}{(-5.9)} & 2.42 \textcolor{green!60!black}{(+0.32)} & 54.5 \textcolor{blue!100!black}{(-10.1)} & 2.35 \textcolor{green!60!black}{(+0.43)}\\
     w/o LG & 60.6 \textcolor{blue!100!black}{(-19.7)} & - & 49.8 \textcolor{blue!100!black}{(-21.3)} & - & 21.2 \textcolor{blue!100!black}{(-37.0)} & - & 19.4 \textcolor{blue!100!black}{(-45.2)} & - \\
    w/o GC \& LG & 55.1 \textcolor{blue!100!black}{(-25.2)} & - & 46.2 \textcolor{blue!100!black}{(-24.9)} & - & 16.0 \textcolor{blue!100!black}{(-42.2)} & - & 15.7 \textcolor{blue!100!black}{(-48.9)} & -  \\

    \bottomrule
  \end{tabular}
  }
  \endgroup
  \label{tab:ablation}
\end{table*}

\noindent \textbf{Generalization Evaluation.} We evaluate generalization along three axes including \textbf{Spatial Generalization}, \textbf{Object Generalization}, and \textbf{Visual Generalization}. As summarized in Table \ref{tab:main}, the Diffusion Policy suffers a large performance drop when confronted with novel objects or light conditions: out-of-distribution (OOD) visual observations substantially degrade its capability. By contrast, GraspVLA shows strong zero-shot generalization and stable performance across evaluation splits, which we attribute to its model capacity and large-scale simulation data. Our method achieves the best results across all three generalization settings, indicating that the graspness cue and grasp pose guidance enable the model to remain robust under OOD visual observations and still generate correct grasp trajectories.

\begin{figure}[t!]
  \centering
  \includegraphics[width=0.47\textwidth]{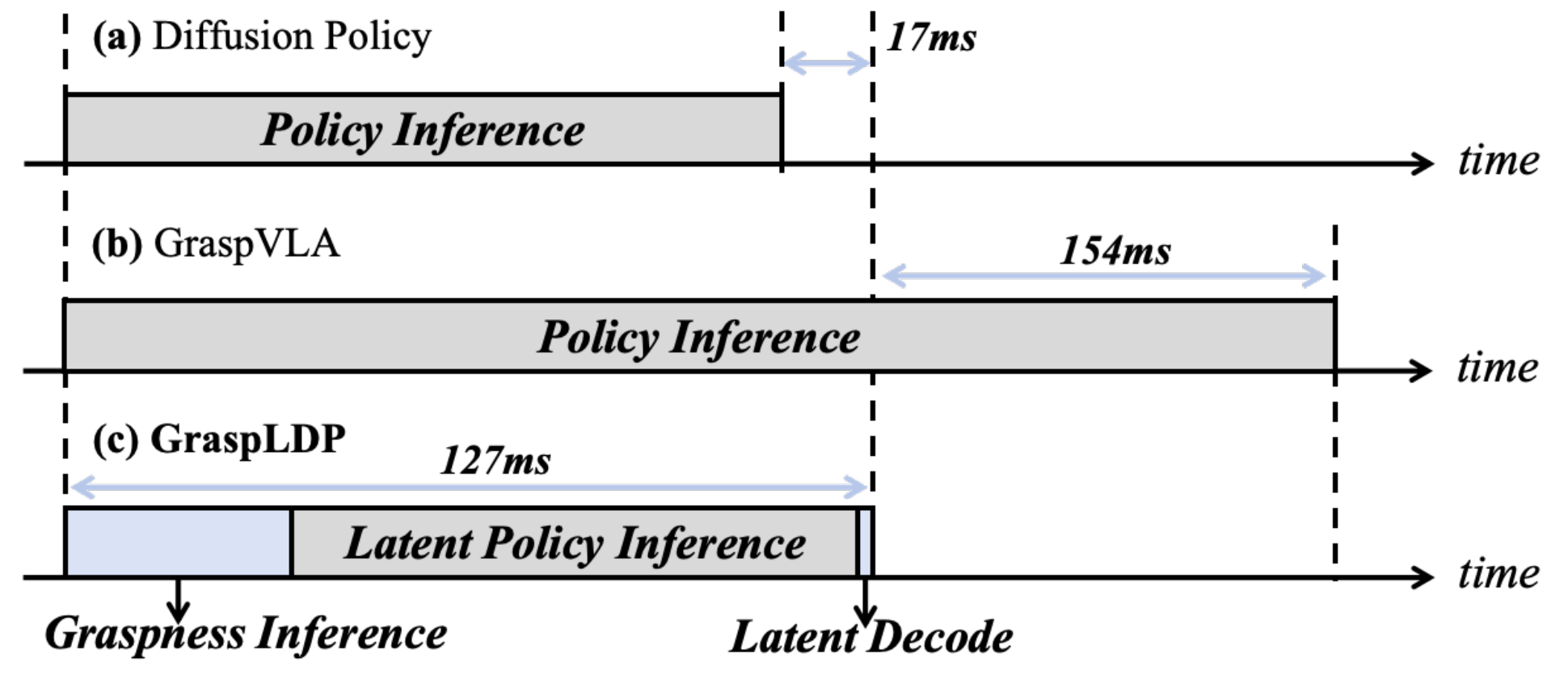}
  \caption{Inference latency of three methods on an RTX 4090 GPU, with the policy action horizon aligned to 8 for each inference. Results of GraspVLA are after acceleration with $torch.compile()$.
}
  \label{fig:infer_time}
\end{figure}
\noindent \textbf{Inference Time Analysis.} Because our method adds extra processing at each inference timestep, a quantitative analysis of inference latency is necessary. As shown in Figure \ref{fig:infer_time}, our method introduces two additional components: graspness inference ($36\ ms$) and latent decode ($<1ms$) compared to vanilla Diffusion Policy. At the same time, a smaller dimensionality of latent space leads to faster policy inference. Overall, GraspLDP is only about 15\% slower than diffusion policy with the same configuration, delivering nearly a twofold improvement in overall success rate. For GraspVLA, its inference latency remains $154ms$ higher than GraspLDP even with acceleration, which means our method can respond faster in dynamic scene.

\subsection{Ablation Study}

\begin{table}[t!]
  \centering
  %\scriptsize
  \footnotesize
  \caption{Ablation study on selection strategy of grasp pose.}
  {\setlength{\tabcolsep}{4pt}%
  \begin{tabular}{m{1.4cm}@{\hskip 6pt}m{1.4cm}@{\hskip 1pt}m{1.4cm}@{\hskip 1pt}m{1.4cm}@{\hskip 1pt}m{1.4cm}@{\hskip 4pt}}
    \toprule
    \textbf{Method} & \centering\textbf{ID} & \centering\textbf{SG} & \centering\textbf{OG} & \ \ \ \ \ \ \ \textbf{VG} \\
    \midrule \midrule
    w/ random & 66.8 \textcolor{blue!100!black}{(-13.5)} & 60.6 \textcolor{blue!100!black}{(-10.5)} & 51.7 \textcolor{blue!100!black}{(-6.5)} & 54.5 \textcolor{blue!100!black}{(-10.1)} \\
    w/ highest & 72.0 \textcolor{blue!100!black}{(-8.3)} & 63.1 \textcolor{blue!100!black}{(-8.0)} & 55.4 \textcolor{blue!100!black}{(-2.8)} & 58.2 \textcolor{blue!100!black}{(-6.4)} \\
    w/ nearest & 69.5 \textcolor{blue!100!black}{(-10.8)} & 61.8 \textcolor{blue!100!black}{(-9.3)} & 53.8 \textcolor{blue!100!black}{(-4.4)} & 56.0 \textcolor{blue!100!black}{(-8.6)} \\
    \midrule
    \rowcolor{gray!20}
    \textbf{w/ HPS} & \textbf{80.3} & \textbf{71.1} & \textbf{58.2} & \textbf{64.6} \\
    \bottomrule
  \end{tabular}
  }
  \label{tab:ablation2}
\end{table}

%We performed extensive ablation studies. 
We first ablate the major components of the framework to evaluate their individual contributions. \textbf{Graspness Cue} and \textbf{Latent Guidance} refer to two incremental augmentations: the former adds the geometry-driven graspness visual cue together with an auxiliary self-supervised reconstruction objective, while the latter conducts grasp pose guidance inside the latent space of the action chunk. We also report \textbf{GFE} to quantify how well each method follows the guidance of target grasp pose.
As shown in Table \ref{tab:ablation}, the Graspness Cue increases overall grasp success rate, with the effect most pronounced in VG evaluation split. We attribute this to the geometric, illumination-invariant nature of graspness: under challenging lighting or visual noise, the graspness cue provides a lighting-robust visual hint that draws the end-effector toward higher graspness regions. Large reduction on SR when we remove LG and increase on GFE when we relpace LG with CG indicate that guiding the policy in the compact latent action representation is a more effective and precise way to incorporate the grasp pose prior. 

Additionally, we perform an ablation study on our proposed grasp pose selection strategy. Using the fully trained GraspLDP model, we replace the HPS at inference time with three simple alternatives: \textbf{random} (uniformly sample a candidate from $\mathcal G$), \textbf{highest} (pick the grasp with the largest grasp score), and \textbf{nearest} (pick the grasp with the smallest SE(3) distance to the current end-effector pose). Table \ref{tab:ablation2} reports the results. The experiments clearly demonstrate the benefit of HPS’s balanced design. HPS jointly trades off grasp quality and kinematic proximity, yielding smoother, more feasible end-effector trajectories and significantly better task success than any single-criterion selection rule.

\subsection{Real World Evaluation}

\begin{figure}[h!]
  \centering
  \includegraphics[width=0.47\textwidth]{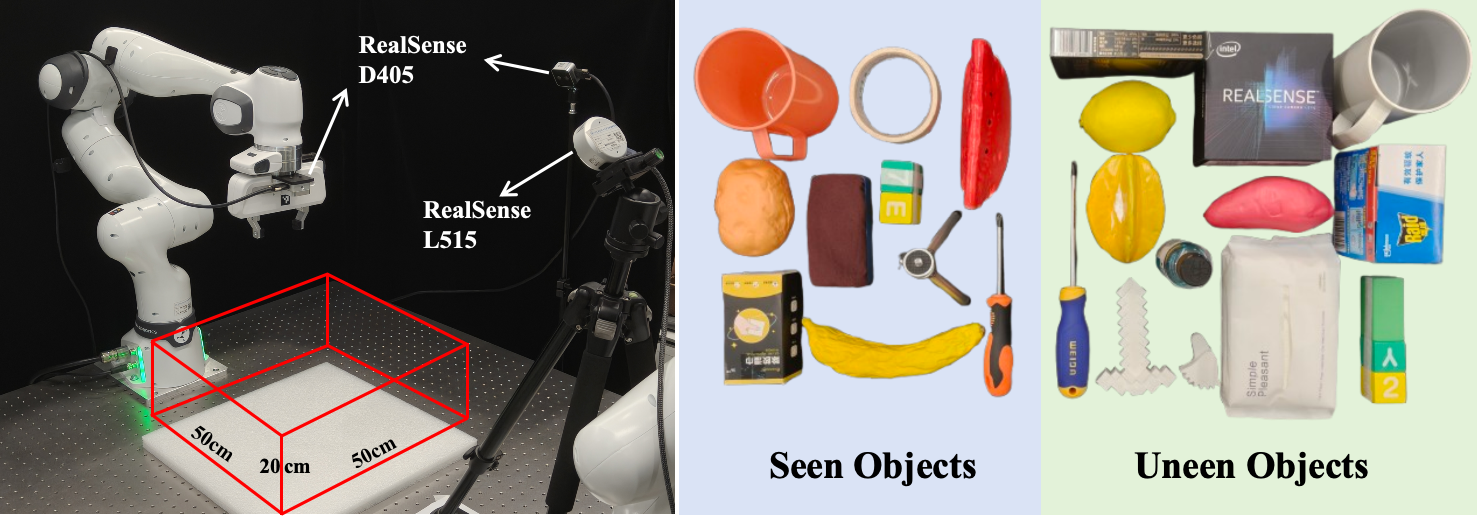}
  \caption{Settings of real world evaluation and daily objects used for training and evaluation.}
  \label{fig:real_world}
\end{figure}

\begin{table}
  \small
  \centering
  \caption{Results of real world evaluation. Because it's difficult to ensure identical object poses in the real world, we merge the ID and SG splits into a single evaluation set.}
  \begin{tabular}{lp{0.9cm}<{\centering}p{0.9cm}<{\centering}p{0.9cm}<{\centering}p{0.9cm}<{\centering}}
    
    \toprule
    \textbf{Method} & \textbf{ID\&SG} & \textbf{OG} & \textbf{VG} & \textbf{Avg}\\
    \midrule \midrule
    Diffusion Policy~\cite{DBLP:conf/rss/ChiFDXCBS23} & 65.0 & 43.0 & 21.0 & 43.0 \\
    GraspVLA~\cite{graspvla} & 29.0 & 27.0 & 22.0 & 26.0 \\
    Anygrasp~\cite{fang23anygrasp} & \underline{82.0} & \textbf{78.0} & \textbf{79.0} & \textbf{79.7} \\
    \rowcolor{gray!20}
    \textbf{GraspLDP} & \textbf{84.0} & \underline{75.0} & \underline{77.0} & \underline{78.7} \\
    \bottomrule
  \end{tabular}
  \label{tab:realworld}
\end{table}

\begin{table*}[ht!]
  \centering
  \small
  \caption{Results of cluttered scenarios evaluation in real world.}
  \begin{tabular}{lm{0.9cm}m{0.9cm}m{0.9cm}m{0.9cm}m{0.9cm}m{0.9cm}m{0.9cm}m{0.9cm}m{0.9cm}m{0.9cm}}
    \toprule
    \multirow{2}{*}{\textbf{Method}} & 
    \multicolumn{2}{c}{\textbf{Scene1}} &\multicolumn{2}{c}{\textbf{Scene2}} &\multicolumn{2}{c}{\textbf{Scene3}} &\multicolumn{2}{c}{\textbf{Scene4}} &\multicolumn{2}{c}{\textbf{Avg}} \\
    \cmidrule(lr){2-3}
    \cmidrule(lr){4-5}
    \cmidrule(lr){6-7}
    \cmidrule(lr){8-9}
    \cmidrule(lr){10-11}&
    \textbf{SR$\uparrow$} & \textbf{SCR$\uparrow$}  & \textbf{SR$\uparrow$} & \textbf{SCR$\uparrow$}  & \textbf{SR$\uparrow$} & \textbf{SCR$\uparrow$}  & \textbf{SR$\uparrow$} & \textbf{SCR$\uparrow$}  & \textbf{SR$\uparrow$} & \textbf{SCR$\uparrow$} \\
    
    \midrule \midrule
    Diffusion Policy~\cite{DBLP:conf/rss/ChiFDXCBS23} & 10.0 & 20.0 & 0.0 & 0.0 & 20.0 & 28.6 & 10.0 & 12.5 & 10.0 & 15.4\\
    GraspVLA~\cite{graspvla} & 20.0 & 40.0 & 0.0 & 0.0 & 30.0 & 42.9 & 10.0 & 12.5 & 15.0 & 23.1\\
    Anygrasp~\cite{fang23anygrasp} & \textbf{83.3} & \textbf{100.0} & 66.7 & \textbf{100.0} & 77.8 & \textbf{100.0} & \textbf{60.0} & \textbf{75.0} & 70.1 & \textbf{92.3} \\
    \rowcolor{gray!20}
    \textbf{GraspLDP} & \textbf{83.3} & \textbf{100.0} & \textbf{100.0} & \textbf{100.0} & \textbf{100.0} & \textbf{100.0} & \textbf{60.0} & \textbf{75.0} & \textbf{82.8} & \textbf{92.3}\\
    \bottomrule
  \end{tabular}
  \label{tab:multi_scene}
\end{table*}
\noindent \textbf{Settings.} Our real world evaluation setup is shown in Figure \ref{fig:real_world}. We perform experiments on a Franka Research 3 manipulator. The wrist-view camera is an Intel RealSense D405, the agent-view camera is an Intel RealSense L515. We include an additional side-view RealSense D405 solely for evaluating GraspVLA. The workspace covers a $50 \times 50 \times 20 \ cm^3$ region. Our object set consists of everyday items: 10 training objects and 13 test  objects, including most rigid bodies and a small number of articulated and deformable objects. For each training object we collect 50 demonstrations (500 demonstrations total) to train both our method and Diffusion Policy; we also compare against GraspVLA and AnyGrasp\cite{fang23anygrasp}. To ensure fair evaluation, the robot’s initial pose, workspace bounds and all camera extrinsics remain fixed across every experiment.
Each evaluation set contains 10 objects, and for each object we run 10 trials with different poses sampled across the workspace.

\begin{figure}
  \centering
  \includegraphics[width=0.44\textwidth]{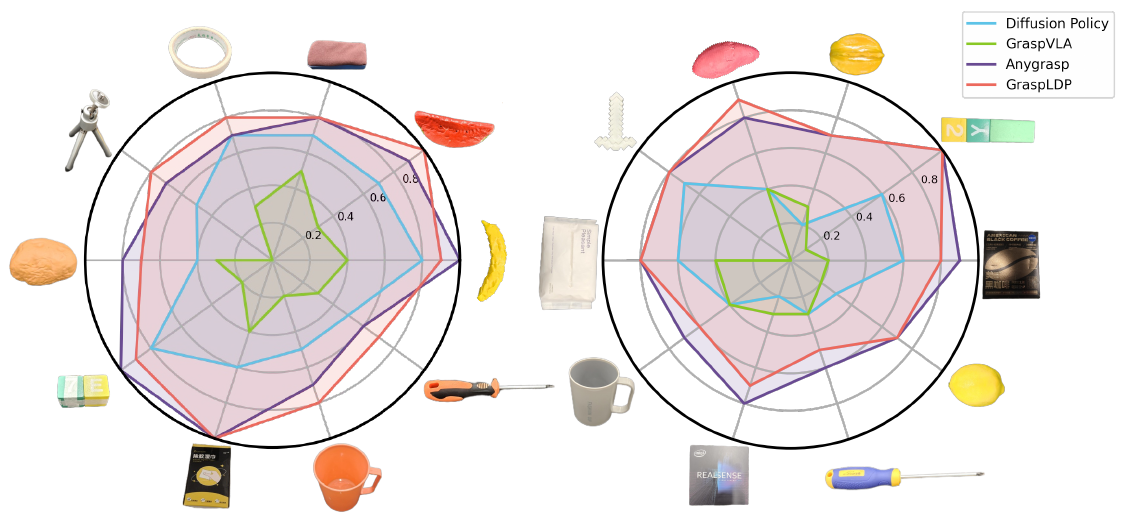}
  \caption{Detailed results of in domain\&spatial generalization and object generalization in the real world.}
  \label{fig:real_world_result}
  \vspace{-0.3cm}
\end{figure}
\noindent \textbf{Main Results.} The overall experimental results are summarized in Table \ref{tab:realworld}. Our method achieves the best success rate of \textbf{84.0\%} on the ID\&SG evaluation, and it maintains strong robustness under novel objects and extreme visual variations---achieving an overall SR comparable to AnyGrasp. This indicates that our design enables the closed-loop policy to inherit the grasp detector’s strong generalization to novel objects and visual changes, significantly outperforming the Diffusion Policy and GraspVLA. Figure \ref{fig:real_world_result} provides a detailed per-object breakdown of each method’s performance for the ID\&SG and OG evaluations.

%\begin{figure}[h!]
%  \centering
%  \includegraphics[width=0.42\textwidth]{fig/multi_scene.pdf}
%  \caption{Initial object placements from cluttered scene 1 to 4.}
%  \label{fig:multi_sence}
%\end{figure}
\noindent \textbf{Cluttered Scenarios Evaluation.}
We conduct grasping experiments in cluttered scenes using the same real robot setup. As shown in Appendix, we design four cluttered scenarios containing from 5 to 8 objects; Scene 4 even includes a more challenging stacked-object configuration. The progressively increasing difficulty makes it clearer to reveal each method’s limits in handling cluttered Scenarios.

The results are summarized in Table \ref{tab:multi_scene}, both our method and AnyGrasp achieve the highest SCR of \textbf{92.3\%}, while our method attains a \textbf{12.7\%} higher SR. This outcome is notable because, unlike AnyGrasp which is trained on multi-object point-cloud data, GraspLDP was trained only on single object grasping demos. GraspLDP successfully grasps every object in Scene 1–3 and still demonstrates strong performance in the stacking scenario, indicating good generalization across the height of workspace which is often overlooked in previous work. Due to zero-shot deployment, GraspVLA could only complete the simplest grasps, though it still outperforms the diffusion policy which is frequently confused by multi-object interference.

\begin{table}[t!]
  \centering
  \small
  \caption{Results of dynamic grasp task.}
  \begin{tabular}{l>{\centering\arraybackslash}m{1.2cm}>{\centering\arraybackslash}m{1.4cm}>{\centering\arraybackslash}m{1.3cm}}
    \toprule
    \textbf{Method} & \textbf{banana\newline moving} & \textbf{watermelon\newline moving} & \textbf{mug\newline handover} \\
    \midrule \midrule
    Diffusion Policy~\cite{DBLP:conf/rss/ChiFDXCBS23} & \xmark/ \xmark/ \xmark & \xmark/ \xmark/ \xmark & \cmark/ \xmark/ \xmark \\
    GraspVLA~\cite{graspvla} & \cmark/ \cmark/ \xmark & \cmark/ \xmark/ \xmark & \xmark/ \xmark/ \xmark \\
    Anygrasp~\cite{fang23anygrasp} & \cmark/ \cmark/ \xmark & \cmark/ \xmark/ \xmark & \cmark/ \xmark/ \xmark \\
    \rowcolor{gray!20}
    \textbf{GraspLDP} & \cmark/ \cmark/ \xmark & \cmark/ \cmark/ \cmark & \cmark/ \cmark/ \xmark \\
    \bottomrule
  \end{tabular}
  \label{tab:dynamic}
\end{table}

\noindent \textbf{Dynamic Grasp.}
We also validate our method’s dynamic grasping capability on the same real robot setup. For GraspLDP and Diffusion Policy we shorten the action horizon from 8 to 4 so the controller can react more promptly to object motion. Results in Table \ref{tab:dynamic} show that the diffusion policy trained on static grasp data almost fails to adapt to dynamic scenes. By contrast, our method, which updates the pose of guiding grasp synchronously, can track, approach, and grasp moving objects. With a higher inference frequency, it achieves markedly better performance than GraspVLA. Compared to AnyGrasp, which uses a tracker to maintain temporal consistency across adjacent frames, rapid changes in the target pose still induce abrupt changes in end-effector trajectories and cause failures, our HPS explicitly accounts for the SE(3) distance between the current end-effector pose and grasp pose candidates, and the policy itself conditions on recent end-effector states. As a result, GraspLDP generates more continuous and smoother grasp trajectories and attains higher success rates.

\section{Conclusion}
\label{sec:conclusion}

In this work, we propose a framework of generalizable latent diffusion policy for grasping. \textbf{GraspLDP} adopt grasp pose guidance in latent action space and leverages auxiliary self-supervised reconstruction of graspness cues to achieve higher grasping precision and improve spatial, visual and object generalization. It can handle cluttered multi-object scenarios and shows promising capability in dynamic scenarios. The method demonstrates strong performance and scalability, offering a promising foundation for future work on robotic foundation model for grasping and manipulation.

\noindent \textbf{Limitations.} Our method may still face challenges when handling highly deformable or fragile objects such as egg or beaker. In future works we plan to incorporate high frequency tactile and force/torque signals  into the framework and explore more general grasping policy.
%\vspace{-0.8cm}
\section*{Acknowledgment}
\label{sec:Acknowledgment}

This work is supported by the National Key Research
and Development Plan (2024YFB3309302).


\renewcommand\thesection{A\arabic{section}}
\setcounter{section}{0}
{
    \small
    \bibliographystyle{ieeenat_fullname}
    \bibliography{main}
}

\clearpage
\setcounter{page}{1}
\maketitlesupplementary

\section{Data Collection for Train and Evaluation}
\textbf{Data efficiency.} 
For simulation experiments, we utilized 12k demonstrations across 20 objects (about 840k frames), while for real-world experiments, we used 500 demonstrations across 10 objects (about 32k frames). Notably, this is substantially less data compared to GraspVLA, which trains on billion-level frames, and is also far smaller than typical VLA models such as $\pi_0$.
We use 12K simulated demonstrations for 20 objects and 500 real-world demonstrations over 10 objects, which is substantially less data than GraspVLA (billion-level frames) and typical VLA models such as $\pi_0$.
Besides, we also evaluated GraspLDP with 1.2K and 120 demonstrations (6 per object), shown in the table below. As training data decreases, our method exhibits significantly better few-shot capabilities than original DP. %Specifically, with only 120 demonstrations, GraspLDP surpasses DP by 29.3\%, demonstrating its effectiveness in few-shot settings.

\begin{table}[h]
  \footnotesize	
  \centering
  \caption{Results of In-domain evaluation with few demonstrations.}
  %\vspace{-4pt}
  \begin{tabular}{@{\hskip 2pt}l@{\hskip 4pt}p{1.5cm}@{\hskip 2pt}p{1.5cm}@{\hskip 2pt}p{1.5cm}@{\hskip 2pt}p{1.5cm}@{\hskip 0pt}}
    \toprule
    \textbf{Method} & \textbf{12K} & \textbf{1.2K} & \textbf{120} & \textbf{Avg}\\
    \midrule
    \midrule
    Diffusion Policy & 62.8 & 41.5 & 13.8 & 39.4 \\
    \textbf{GraspLDP} & \textbf{80.3} \textcolor{green!60!black}{(+17.5)} & \textbf{64.6} \textcolor{green!60!black}{(+23.1)} & \textbf{43.1} \textcolor{green!60!black}{(+29.3)} & \textbf{62.7} \textcolor{green!60!black}{(+23.3)} \\
    \bottomrule
  \end{tabular}
  \label{tab:few-shot}
\end{table}

\noindent \textbf{Using different grasp detectors.} We evaluated several grasp detectors with varying performance levels as priors, as shown in the table below. Our method consistently outperforms the grasp detection baseline with traditional motion planning. Notably, the performance gain is even more significant when the initial grasp poses are \textbf{suboptimal}, demonstrating the robustness and corrective capability of our proposed closed-loop diffusion-based policy. 
%We will include this in the revised version.

% We switch between different grasp detectors to provide guiding grasp poses, and additionally apply randomization to the grasp poses predicted by GSNet to simulate scenarios with lower grasp pose quality. %As shown in Figure \ref{tab:grasp},%
% When the quality of the guiding grasp poses degrades, our method exhibits stronger robustness and demonstrates clear advantages over motion planners.
\begin{table}[h]
  %\small
  \footnotesize	
  \centering
  %\vspace{-2pt}
  \caption{Results of In-domain setting with different grasp detector.}
  \begin{tabular}{@{\hskip 4pt}lm{1.5cm}@{\hskip 2pt}m{1.5cm}@{\hskip 2pt}m{1.5cm}@{\hskip 0pt}m{2.2cm}}
    \toprule
    \textbf{Method} & \textbf{GSNet} & \textbf{SBG} & \textbf{GraspNet} \\
    \midrule 
    \midrule
    Grasp Detection & 78.5 & 73.8 & 70.8 \\
    \textbf{GraspLDP} & \textbf{80.3} \textcolor{green!60!black}{(+1.8)} &  \textbf{76.9} \textcolor{green!60!black}{(+3.1)} &  \textbf{75.4} \textcolor{green!60!black}{(+4.6)} \\
    \bottomrule
  \end{tabular}
  \label{tab:grasp}
\end{table}

\section{Data Collection for Train and Evaluation}
\begin{figure}[h!]
  \centering
  \includegraphics[width=0.98\linewidth]{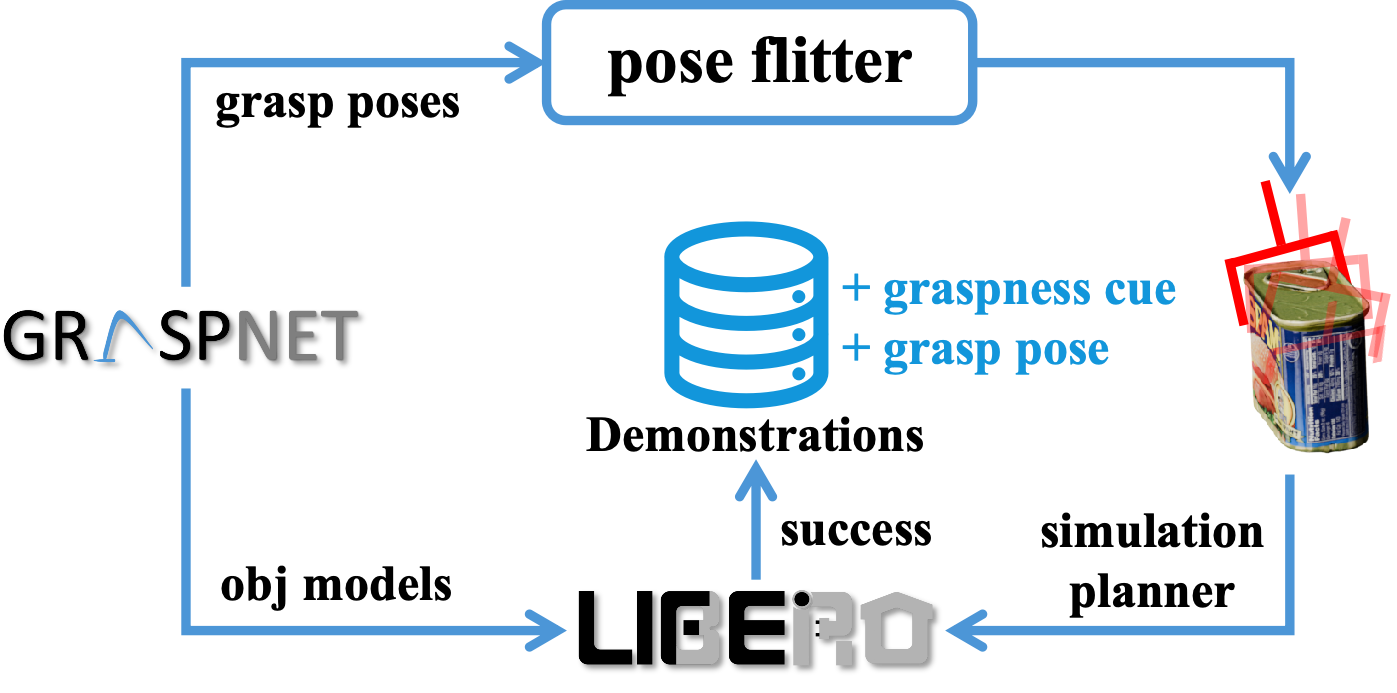}
  \caption{Illustration of the data collection pipeline in simulator.}
  \label{fig:data}
\end{figure}

There exist many open-source datasets of expert demonstrations, collected on real robots or in simulator, that are widely used to train imitation learning policies. However, these datasets are often heterogeneous and contain relatively few demonstrations specifically targeted at grasping tasks; moreover, interaction patterns with objects tend to be narrow (for instance, grasp poses are commonly concentrated around simple top-down planar grasps). Such distributional biases strongly limit policy performance on diverse or precise grasping tasks.

To address these limitation, we redesigned our demonstration collection pipeline starting from a grasp detection dataset. GraspNet-1Billion~\cite{fang2020graspnet} provides dense grasp annotations for 78 objects (each object appearing in multiple poses and scenes), which nicely matches our needs. As illustrated in Figure \ref{fig:data}, we import GraspNet-1Billion’s 3D object assets into the LIBERO~\cite{liu2023libero} benchmark (built on the robosuite~\cite{zhu2020robosuite} simulator, which in turn uses MuJoCo~\cite{todorov2012mujoco} as physics engine). After correct transformation, we can place any object in any training scene in LIBERO and directly sample candidate grasp poses from the dataset labels.
For each object we sample a large set of high-quality grasp poses and apply NMS to avoid clustering poses in a small region and to ensure diversity. We generate end-effector motion trajectories by Spherical Linear Interpolation(SLERP) for rotations and Linear Interpolation for translations. We then simulate these trajectories in LIBERO and, at each timestep, record the wrist-view RGB images augmented with the graspness visual cue. We log the target grasp pose per episode and include only demonstrations that succeed in simulation into our final dataset. We employ a heuristic speed de-biasing procedure during trajectory synthesis so that the resulting trajectories share as uniform velocities as possible cause prior study~\cite{shi2025diversity} have shown that distributional shifts in trajectory speeds across datasets increase the difficulty of policy learning.

\section{Details of Real World Evaluation}
\subsection{Language Label for Objects}
\begin{figure}[ht!]
  \centering
  \includegraphics[width=0.98\linewidth]{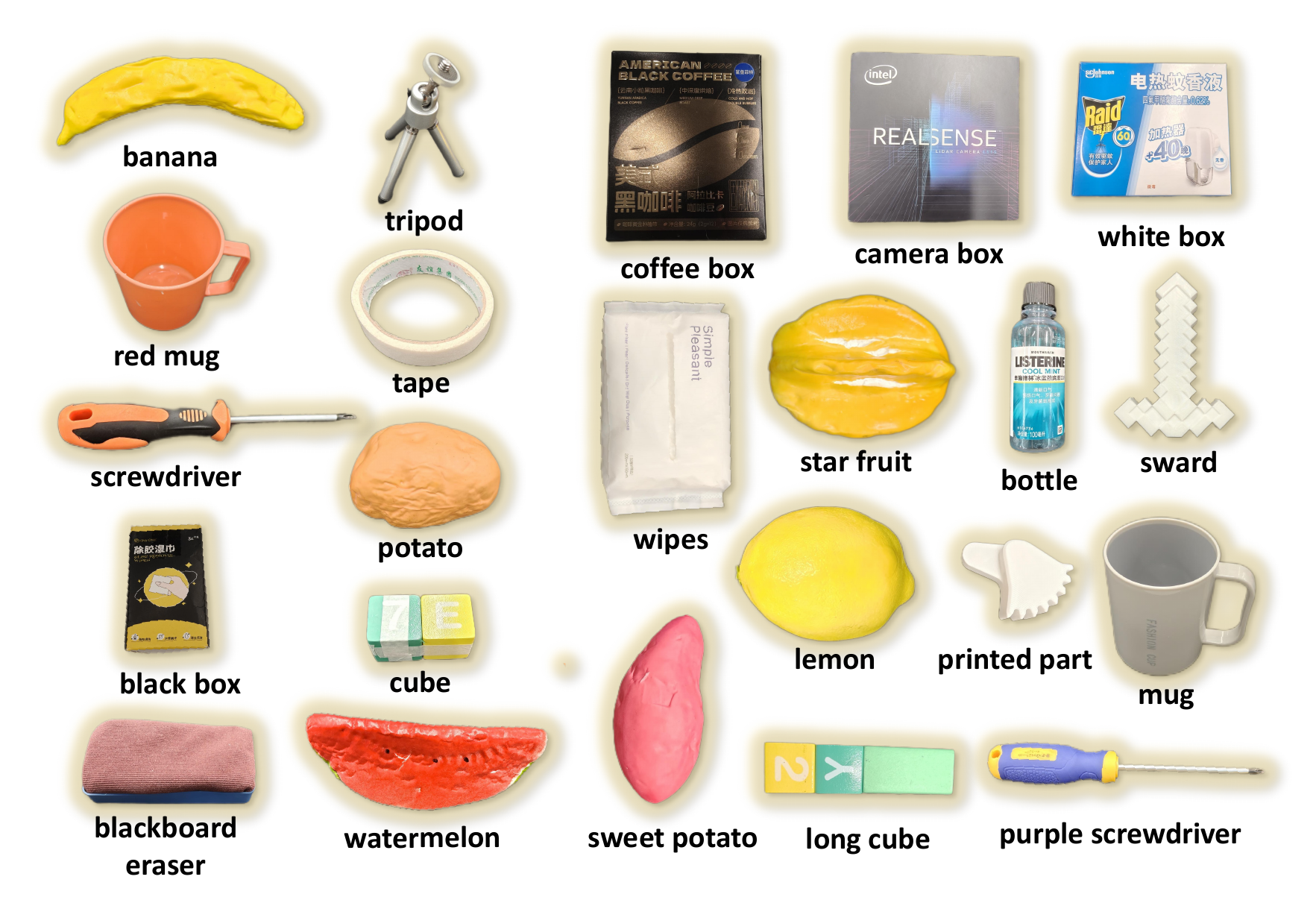}
  \caption{Language label for each object.}
  \label{fig:objects}
\end{figure}
Figure \ref{fig:objects} shows the language labels of objects used in our real world evaluation. These labels are primarily used for evaluating the GraspVLA\cite{graspvla} and are also employed within the cluttered scenarios evaluation.

\subsection{Cluttered Scenarios Evaluation}
\begin{figure}[ht!]
  \centering
  \includegraphics[width=0.98\linewidth]{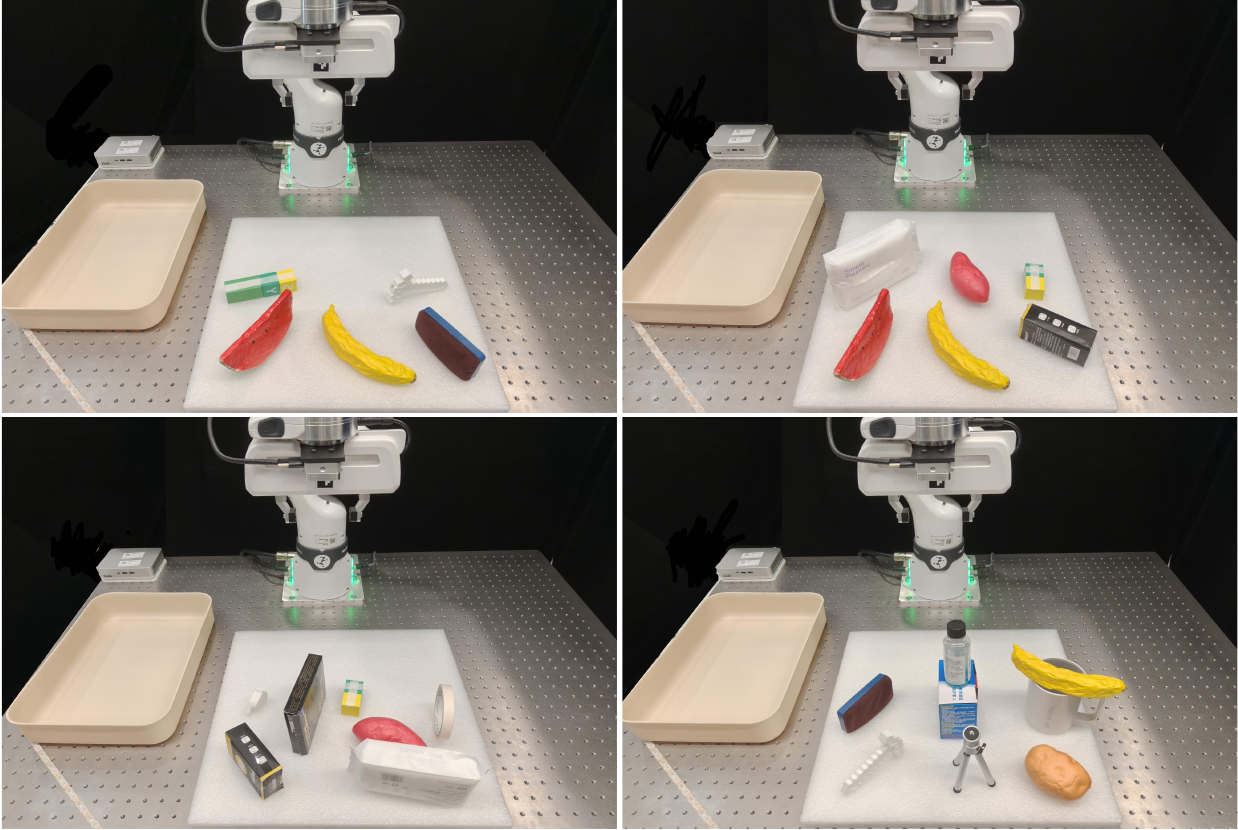}
  \caption{Initial object placements in cluttered scenerios evaluation from scene 1 to scene 4.}
  \label{fig:multi_scene}
\end{figure}
As shown in Figure \ref{fig:multi_scene}, we set up four cluttered scenes, each containing both seen and unseen objects. The number of objects increases from 5 to 8 across scenes 1–4, and scene 4 includes a particularly challenging stacked-object configuration. The goal of task is to grasp and clear all objects from the tabletop; each scene allows up to 10 grasp attempts. For GraspVLA we form language instructions using each object’s label (\eg  \textit{“pick up the banana”}). For Anygrasp\cite{fang23anygrasp} and our method, we follow an approach that uses Grounding DINO\cite{liu2024grounding}, a robust open-vocabulary object detector, to filter candidate grasp poses for the target object. During trials, when the timestep reaches a predefined threshold we invoke motion planning to move the gripper to a fixed waypoint and open it to complete the placement step, thereby ensuring continuity of the table-clearing process.
\label{sec:rationale}
\begin{table}[t!]
  \centering
  %\small
  \caption{Training and inference hyperparameters of Action Latent Learning.}
  \begin{tabular}{cc}
    \toprule
    \textbf{Parameter} & \textbf{Value} \\
    \midrule
    \midrule
    horizon & 16 \\
    n\_obs\_steps & 2 \\
    n\_latent\_dims & 16 \\
    use\_conv\_encoder & True \\
    conv\_latent\_dims & 64 \\
    conv\_layer\_num & 1 \\
    use\_rnn\_decoder & True \\
    rnn\_latent\_dims & 64 \\
    rnn\_layer\_num & 1 \\
    kl\_multiplier & 1e-6 \\ 
    n\_embed & 16 \\
    use\_vq & False \\
    dataloader.batch\_size & 128 \\
    dataloader.num\_workers & 4 \\
    optimizer.\_target\_ & \texttt{torch.optim.AdamW} \\
    optimizer.lr & $1.0\times 10^{-3}$ \\
    optimizer.weight\_decay & $1.0\times 10^{-4}$ \\
    training.lr\_scheduler & cosine \\
    training.lr\_warmup\_steps & 100 \\
    training.num\_epochs & 1000 \\
    \bottomrule
  \end{tabular}
  \label{tab:hyperparams1}
\end{table}
\subsection{Description of Dynamic Grasp Tasks}
For moving target objects, the policy must continuously update its visual observations and the guidance of grasp pose to realize closed-loop dynamic grasps. In the \textit{banana moving} and \textit{watermelon moving} tasks, the object starts from a corner of the workspace and traverses the entire area; in the \textit{mug handover} task, a human holds the mug by the handle and moves it horizontally and vertically with the goal of having the gripper catch the wall of mug and complete the handover. Because we moved the objects manually, it is difficult to guarantee that the movements are exactly the same across different methods. Therefore, to demonstrate advantage of GraspLDP, we always ensured that the object movement distance was the longest in the evaluation of our method.

\begin{table}[h]
  \centering
  %\small
  \caption{Training and inference hyperparameters  of Diffusion on Latent Action Space.}
  \begin{tabular}{@{\hskip 2pt}c@{\hskip 6pt}c@{\hskip 2pt}}
    \toprule
    \textbf{Parameter} & \textbf{Value} \\
    \midrule
    \midrule
    observation\_horizon & 2 \\
    prediction\_horizon & 16 \\
    action\_horizon & 8 \\
    unet.diffusion\_step\_embed\_dim & 256 \\
    unet.down\_dims & [256,512,1024] \\
    unet.kernel\_size & 5 \\
    unet.n\_groups & 8 \\
    enable\_ddim & True \\
    num\_training\_timesteps & 100 \\
    num\_inference\_timesteps & 10 \\
    prediction\_type & epsilon \\
    use\_recon & True \\
    recon\_loss\_weight & 0.2 \\
    dataloader.batch\_size & 64 \\
    dataloader.num\_workers & 8 \\
    optimizer.\_target\_ & \texttt{torch.optim.AdamW} \\
    optimizer.lr & $1.0\times 10^{-4}$ \\
    optimizer.weight\_decay & $1.0\times 10^{-6}$ \\
    training.lr\_scheduler & cosine \\
    training.epoch\_every\_n\_steps & 100 \\
    training.num\_epochs & 1500 \\
    training.use\_ema & True \\
    training.ema\_power & 0.75 \\
    training.ema\_power & 0.75 \\

    \bottomrule
  \end{tabular}
  \label{tab:hyperparams2}
\end{table}

\section{Model Implementation}
In Action Latent Learning, we use a 1D-CNN encoder $\mathcal{E}$ as action chunk encoder, and adopt a GRU as action chunk decoder $\mathcal{D}$. During  training of the VAE, we set $\lambda = 1\times 10^{-6}$. Keeping it as such a small value reduces the strength of KL regularization and improves reconstruction quality, since we do not need to sample directly from this latent space for generation. More details of the VAE model and training settings are show in Table \ref{tab:hyperparams1}.

In the Diffusion on Latent Action Space, we freeze the parameters of VAE and use the action latent produced by the action chunk encoder as the supervision of denoising target. We use ResNet18 as our obs encoder to process agent-view and wrist-view image with size of $H=W=256$. We use DDIM~\cite{songdenoising} as the noise scheduler, and the number of denoising steps is 100 during training and 10 at inference. For the graspness cue, we set $\tau = 0.2$ to strike a balance between sufficiently covering graspalbe regions and introduing less noise. During auxiliary reconstruction of the graspness cue, we follow prior work~\cite{li2024crossway} that introduces self-supervised reconstruction into diffusion policy and set the reconstruction weight $\lambda_{\mathrm{Recon.}} = 0.2$ after several experiments. All rotations are represented with the 6D rotation representation~\cite{zhou2019continuity}, which provides better continuity in the numerical and 3D rotation spaces for neural network learning. More details of the latent diffusion model and training settings are show in Table \ref{tab:hyperparams2}.

In Heuristic Pose Selector (HPS) of inference pipeline, a pre-trained GSNet~\cite{Graspness}, the core grasp detection module of AnyGrasp~\cite{fang23anygrasp}, are used as grasp detector. We set $k = 30$ to filter low scored grasp candidates. And for matrix $W$, we set $w_t = 100 $ and $ w_r = 20$ to achieve a trade-off between measurement of translation and rotation following \cite{goyal2022ifor}.

In our real world experiments with both Diffusion Policy and GraspLDP, inference latency commonly induces jitter when switching action chunks, which in turn perturbs grasp trajectories. To mitigate this effect, we simply discard the first three actions predicted by the policy at each inference process. This pragmatic process substantially reduces transition-induced oscillations and yields noticeably smoother and more reliable grasp executions.

% 
%Having the supplementary compiled together with the main paper means that:
% 
%\begin{itemize}
% 
%To split the supplementary pages from the main paper, you can use \href{https://support.apple.com/en-ca/guide/preview/prvw11793/mac#:~:text=Delete%20a%20page%20from%20a,or%20choose%20Edit%20%3E%20Delete).}{Preview (on macOS)}, \href{https://www.adobe.com/acrobat/how-to/delete-pages-from-pdf.html#:~:text=Choose%20%E2%80%9CTools%E2%80%9D%20%3E%20%E2%80%9COrganize,or%20pages%20from%20the%20file.}{Adobe Acrobat} (on all OSs), as well as \href{https://superuser.com/questions/517986/is-it-possible-to-delete-some-pages-of-a-pdf-document}{command line tools}.

%\section{Visualization of Real World Evaluation}
%We attach a video in supplementary material to visualize our real world evaluation in each evaluation setting and demonstrate the effectiveness of our proposed method by comparing with other baseline method.
\end{document}